# EGI: A Multimodal Emotional AI Framework for Enhancing Scrum Master Real-time Self-Awareness

Jingni Huang†
Department of Computer Science
University of Oxford
Oxford, United Kingdom
jingni.huang@kellogg.ox.ac.uk, jingnih@gmail.com

Peter Bloodsworth
Department of Computer Science
University of Oxford
Oxford, United Kingdom
peter.bloodsworth@cs.ox.ac.uk

## ABSTRACT

While increasing research focuses on the emotional well-being of agile team members, a significant gap remains in emotion monitoring studies for Scrum Masters and meeting organizers, whose impact on team dynamics is crucial. This paper proposes a novel application integrating four carefully selected and recommended AI models to monitor the unconsciously expressed emotions of these key roles. This is achieved through: real- time transcription using a speech-to-text model; thresholding for intonation analysis to detect emotional cues in prosody; applying emotion-based vocabulary matching to identify sentiment in spoken content; and providing context-aware suggestions containing emotion keywords using an open-source, multi-module AI API. The system achieved an ASR word error rate WER of 10% in simulated meeting environments. Our evaluation shows that real-time feedback significantly improves emotion awareness during simulated agile meetings, providing Scrum Masters and meeting organizers with real-time and practical suggestions to help them quickly identify and minimize the expression of negative emotions, fostering more positive and effective team interactions.

## 1 Introduction

Agile software development [1] emphasizes effective communication at the heart of project success[2-6]. However, the iterative and fast-paced nature of agile often leads to team stress and emotional friction. While existing research extensively examines the collective emotional response of agile teams [47], a significant research gap remains: the emotional monitoring of the Scrum Master. As the enforcer of Scrum values, a Scrum Master's emotional stability is critical to sustaining team morale and productivity [7]. An unaddressed display of anxiety or frustration by a leader can lead to a loss of confidence among engineers and decreased productivity.

Current AI solutions often struggle with the nuanced, real-world emotional states found in professional settings. While large-scale language models excel in text processing, they frequently lack the multi-modal integration required to provide timely, actionable feedback during live interactions. Our research addresses this by introducing the EGI. Unlike existing conversational tools, EGI employs a multimodal approach that fuses automatic speech recognition (ASR [8]) with real-time prosodic analysis. By combining "what" is said (linguistic content) with "how" it is said (intonation), EGI provides Scrum Masters with immediate visual and textual feedback, facilitating self-awareness and fostering a positive team environment without the intrusive need for video-based monitoring.

## 2 Methodology

Our approach integrates Human-Computer Interaction (HCI) design principles with rigorous signal processing to meet the low-latency requirements of real-time agile ceremonies.

To ensure high-performance speech-to-text (STT) integration, we conducted an extensive evaluation of deep learning architectures, including RNN, LSTM, and Transformer-based models. We specifically compared Wav2Vec 2.0 and OpenAI's Whisper model. Our experimental validation using established datasets, such as the Speech Emotion Recognition Voice Dataset [44] and the Opinions Lexicon [45], demonstrated that while Wav2Vec offers low latency, Whisper provides superior Word Error Rate (WER) performance in noisy, multi-accent meeting environments. Consequently, Whisper was selected for the linguistic pipeline.

## 3 Application System Architect

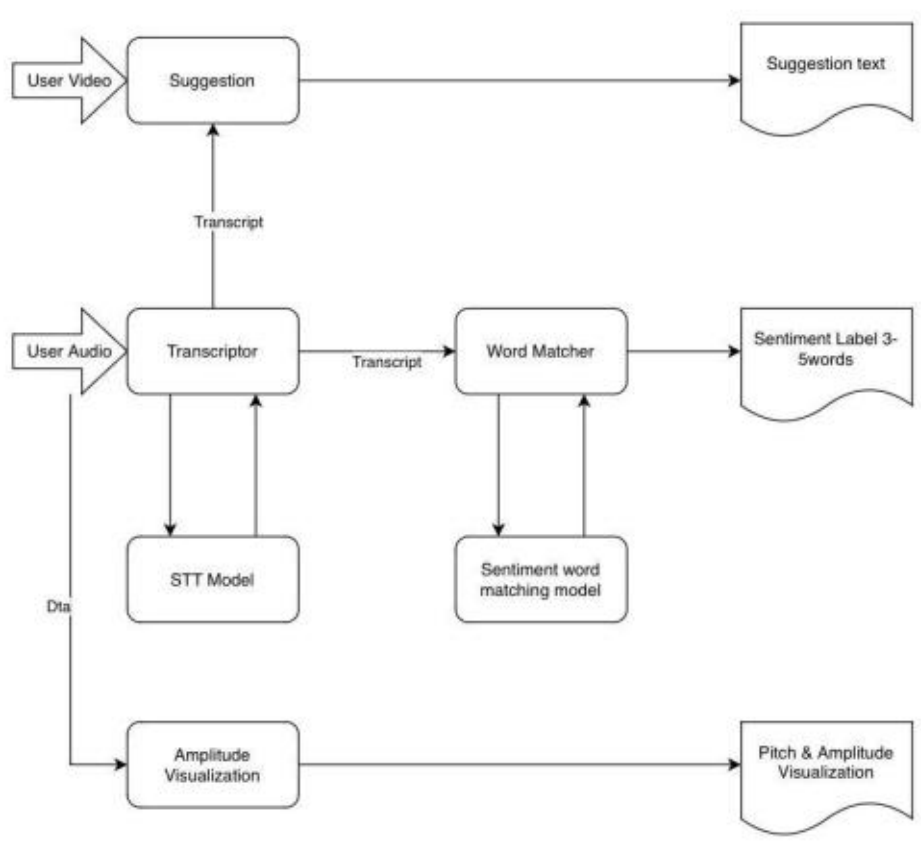

Figure 1: System Architecture(User Video track is future work)

From figure 1 system architecture we could see the application is divided into 4 functional models. First model speech-to- text: Speech-to-Text: This is the first step, using ASR to get the transcript text to convert spoken words into text, laying the foundation for subsequent analysis.

### 3.1 Model Evaluation and Selection

Second mode intonation visualization: Decode the original audio to plot pitch and Amplitude for visualization as the figure 1 Demo's first and second dynamics visualization. Visualization can immediately grab the eye's attention and form a sharp contrast with the continuous waveform of voice intonation, with sudden high and low pitches.

Third model emotion keyword prompt: Transcript text from first model speech to text as input for matching the sentiment word library of positive and negative sentiment words, then display the result on the keyword display bar. It only shows 3- 5 words which reduce reading time and improve first glance recognition under the intonation visualization.

Fourth model suggestions: This is an optional enhancement module. Using the first model text output for combining with multiModal model gpt-4o-transcribe module open source in the end for more complex sentiment analysis and response suggestions, if the scrum master requests extra support.

The system architecture separates core real-time feedback functionality (tone visualization and short keywords) from complex on-demand analysis (recommendations). Because the application requires only microphone input, not a camera, this design ensures real-time performance while balancing cost- effectiveness and user comfort. Advanced multimodal options are also available for advanced users. The ultimate goal of the entire system is to help Scrum Masters quickly analyze their sentiment (positive or negative) and provide immediate response recommendations without increasing the burden on the Scrum Master.

### 3.3 Signal Processing and Data Flow

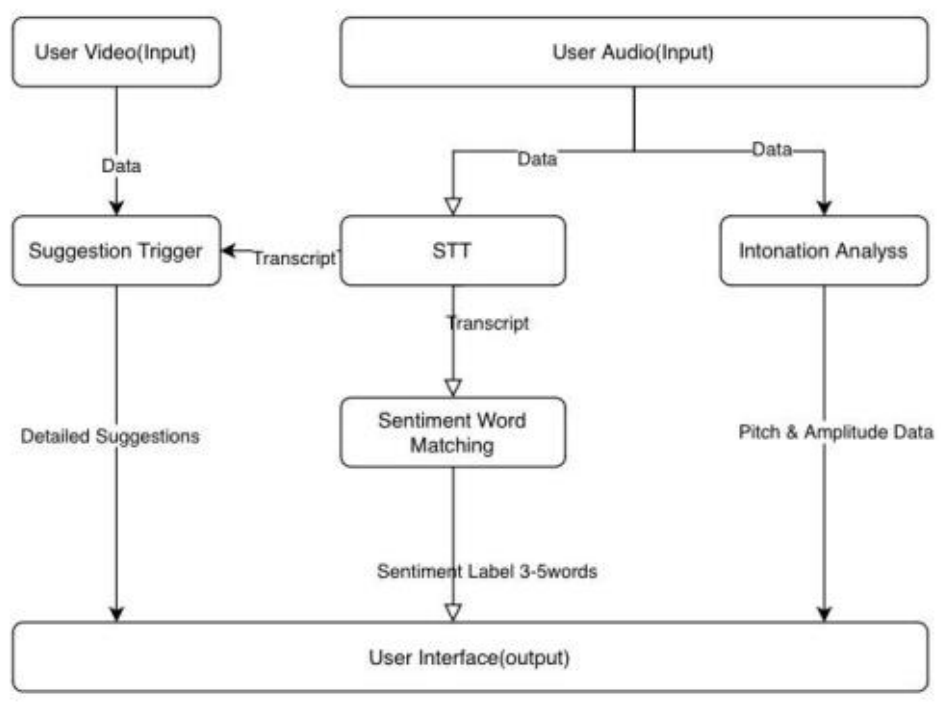

Figure 2: Application Data Flow Diagram (User Video track is future work)

The EGI framework optimizes the inference pipeline through a decoupled, resource- aware task allocation strategy. As illustrated in the data flow (Figure 2), the system processes multimodal inputs via two primary parallel paths to capture a holistic emotional profile:

Path A (Linguistic Path): This path utilizes the Whisper model to generate high-fidelity transcripts from raw audio. These transcripts are then passed to the GPT-4o API for advanced semantic inference and the generation of context-aware suggestions. This ensures that the system leverages the superior reasoning capabilities of Large Multimodal Models (LMMs) for complex emotional support.

Path B (Acoustic Path): Simultaneously, this path performs local Digital Signal Processing (DSP) using libraries such as librosa or Aubio. It extracts key prosodic features, specifically the fundamental frequency (F0) for pitch and root mean square (RMS) energy for amplitude. By processing these features locally, the system ensures the delivery of time-sensitive emotional cues (intonation) with zero network latency**, providing a responsive and fluid user experience.

This dual-path strategy enables EGI to capture emotional valence that pure text analysis typically misses-such as the discrepancy between a positive transcript and a negative tone. Furthermore, the EGI framework is designed for scalability. While the current implementation prioritizes acoustic and linguistic modalities to maintain low latency and user privacy, the architecture reserves a User Video track (as shown in Figure 1) for future integration of facial expression analysis, which will further enrich the multimodal emotional profile.

# 4 Experimental Evaluation

## 4.1 Model Selection and Benchmarking

Table 3: Wav2Vec 2.0 with Whisper Comparison[30]

| Dataset | wav2vec 2.0 Large (no LM) | Whisper Large V2 |
|---|---|---|
| LibriSpeech Clean | **2.7** | **2.7** |
| Artie | 24.5 | **6.2** |
| Common Voice | 29.9 | **9.0** |
| Fleurs En | 14.6 | **4.4** |
| Tedlium | 10.5 | **4.0** |
| CHiME6 | 65.8 | **25.5** |
| VoxPopuli En | 17.9 | **7.3** |
| CORAAL | 35.6 | **16.2** |
| AMI IHM | 37.0 | **16.9** |
| Switchboard | 28.3 | **13.8** |
| CallHome | 34.8 | **17.6** |
| WSJ | 7.7 | **3.9** |
| AMI SDM1 | 67.6 | **36.4** |
| LibriSpeech Other | 6.2 | **5.2** |
| Average | 29.3 | **12.8** |

According to literature benchmarks [30][44][45], Whisper demonstrates superior robustness compared to Wav2Vec 2.0 in handling diverse accents and noisy environments typical of agile meetings. While Wav2Vec 2.0 offers slightly faster inference as denoted as table 3, we selected Whisper to prioritize the textual accuracy required for reliable downstream sentiment analysis.

## 4.2 Databases

Speech Emotion Recognition Dataset [44]: We chose this concise dataset because although it contains only 68 audio samples, it contains accurately labeled and transcribed text, and the audio contains four different emotional expressions, which fully meets the database requirements for our initial experiments.

These texts are read aloud in English representing four different emotional states: euphoria, joy, sadness, and surprise. Each audio clip captures the speaker‘s tone, intonation, and nuances in expressing emotions through sound. This dataset was chosen because the audio clips themselves capture the different intonations, nuances, and subtleties of emotional speech, making them ideal for validating the robustness of our system in unpredictable and emotional environments such as agile meetings. The name of each audio clip itself provides a keyword for true emotion.

Opinion Lexicon[45]: A sentiment lexicon containing 2006 positive and 4783 negative English words, used in keyword suggestion models to identify sentiment words in transcribed text.

## 4.3 Experimental Results and Discussion

Our end-to-end evaluation aimed to demonstrate the effectiveness of an ensemble of four models using a pre-prepared dataset. To this end, we leveraged two key datasets: the Speech Emotion Recognition Dataset and the Opinion Dictionary. The former contains audio clips categorized by four distinct emotions, providing the foundational data for our intonation and ASR experiments. The latter is the core of our emotion tagging model.

The performance of the speech-to-text module is evaluated using the Word Error Rate (WER), defined as:

$$WER = (S + D + I) / N \quad (1)$$

Where

S is the number of substitutions,

D is deletions,

I is insertions,

N is the total number of words in the reference transcript.

Keep leanings and potential optimization include:

1. Implementing noise reduction techniques to achieve strong real-time performance;

2. Practice prompt engineering to optimize the output of the AI suggestion model.

Table 3: models performance and comparison

| Model | Approach | Accuracy/Result | Key Challenges | Proposed Optimizations |
|---|---|---|---|---|
| ASR | Speech-to-text Whisper transcription | WER: 10% | Small dataset with varied intonations and noisy; high WER | pre-process noise, refine kenLM integration |
| Intonation visualization | Waveform and pitch visualization | Qualitative pitch-emotion correlation | Subjective interpretation of pitch levels | Threshold as reference |
| keyword prompt | Text matching with opinion lexicon | String prompts | Limited to lexical sentiment; misses contextual tone | Integrate Intonation visualization model to text dataset; NLP models (e.g., BERT) for nuanced sentiment |
| AI Suggestion Generator | GPT-4o for further feedback | Context-aware responses | Dependency on external APIs; latency | Cache frequent queries; fine-tune local LLM for emotion-specific suggestions |

As table 3 models performance and comparison shows the models experiment comparison:

1. Speech-to-Text (ASR): Whisper

This experiment uses the Whisper speech- to-text model and decodes it to accelerate processing.Preliminary results show a word error rate (WER) of 10%. WER is a key metric for measuring the performance of speech recognition systems. A 10% WER means that 10 out of every 100 words are misrecognized, inserted, or deleted. This result indicates that while the model works, its accuracy still has significant room for improvement.

Major Challenges: The primary challenge stems from the dataset's limitations. While the dataset was chosen to ensure model robustness, its small size and wide range of intonations and background noise significantly interfere with the model's recognition accuracy. The small dataset limits the model's learning and generalization capabilities, while the varying intonations and noise increase the difficulty of recognition. The high WER is a direct reflection of these challenges.

Optimization Solutions: To address these challenges, the experiment proposed two main optimization measures. Preprocessing Noise: By reducing the audio noise, we can provide the model with cleaner input, thereby improving its recognition accuracy. Optimizing KenLM Integration: KenLM is an efficient language model toolkit used to improve decoding results in speech recognition.

2. Intonation Visualization: Waveform and Pitch Visualization

This section aims to visualize the waveform and pitch of speech to qualitatively analyze the relationship between pitch and emotion. By converting audio data into waveform graphs and pitch curves, researchers can visually observe pitch variations and subjectively infer their correlation with the speaker's emotion. For example, a sharp rise in pitch may indicate surprise or anger.

Key Challenge: A core challenge of this approach lies in the subjective nature of the pitch-emotion association. While pitch can provide emotional clues, its interpretation is highly dependent on the speaker's cultural background, context, and individual perception. Objective and accurate emotional judgment based solely on pitch level is difficult, as the same emotion (such as excitement) can manifest differently in different speakers or contexts.

Optimization Solution: To address this subjectivity, the project proposes a simple optimization method: setting a threshold as a reference. For example, a baseline pitch value can be set, and any pitch variation exceeding this threshold is considered emotionally expressive, providing a relatively objective starting point for qualitative analysis.

3. Sentiment keywording: Text Matching and Sentiment Dictionaries

To assign sentiment keywords to text content, the project employed a text matching method based on sentiment dictionaries. This method matches the input transcription against a predefined list of sentiment terms. For example, text containing words like "good" and "like" might be keyworded as positive. This method can complement the tone visualization model described above. When changes in speech tone are insufficient to convey emotion, the context of the conversation can be further referenced.

Main Challenge: This method has significant limitations. It is limited to sentiment analysis at the lexical level and ignores context. Many words are inherently neutral but can take on strong emotional overtones in specific contexts. Furthermore, complex linguistic phenomena such as irony and puns cannot be identified through simple lexicon matching, resulting in the analysis results failing to capture the true emotional tone of the text. However, future research will focus on this aspect and further data testing in conjunction with the tone visualization model described above.

Optimization: To improve the accuracy and depth of sentiment analysis, the project recommends integrating context-aware natural language processing (NLP) models, such as BERT. And as mentioned earlier, it is combined with the previous intonation visualization to perform data testing and parameter optimization.

4. AI Recommendation Generation: Using GPT-4o

This project leverages the powerful generative capabilities of GPT-4o to provide contextual recommendations. GPT-4o is a state-of-the-art large language model (LLM) that can understand and generate high- quality, contextually relevant text. By taking previously transcribed text and camera- projected video as input, GPT-4o can provide more insightful and targeted feedback or recommendations.

Key Challenges: The primary challenge with this approach lies in its reliance on external APIs and the resulting latency. Each call to GPT-4o requires a request to an external server, which not only incurs costs but can also slow down application response times due to network latency and server response times. For applications requiring real-time feedback, this latency needs to be further improved.

Optimization: Caching Frequent Queries: For common or repeated queries, the results can be cached. When the same query is encountered again, the answer can be retrieved directly from the local cache without making another call to the external API. This can significantly reduce latency and API call costs.

Fine-tuning a local LLM for specific emotions: This project proposes using a smaller, more lightweight local large language model and fine-tuning it for specific emotion-related recommendations. While the local LLM may not be as general as GPT-4o, through specialized training, it can perform well on specific tasks while avoiding all the problems of external APIs and achieving fast and low-cost inference.

The ultimate goal of these experiments is to verify that all components meet their established performance targets - such as achieving a relative low WER and maintaining real-time processing latency - thereby confirming the overall system's practical efficacy.

## 5 Demo Interface

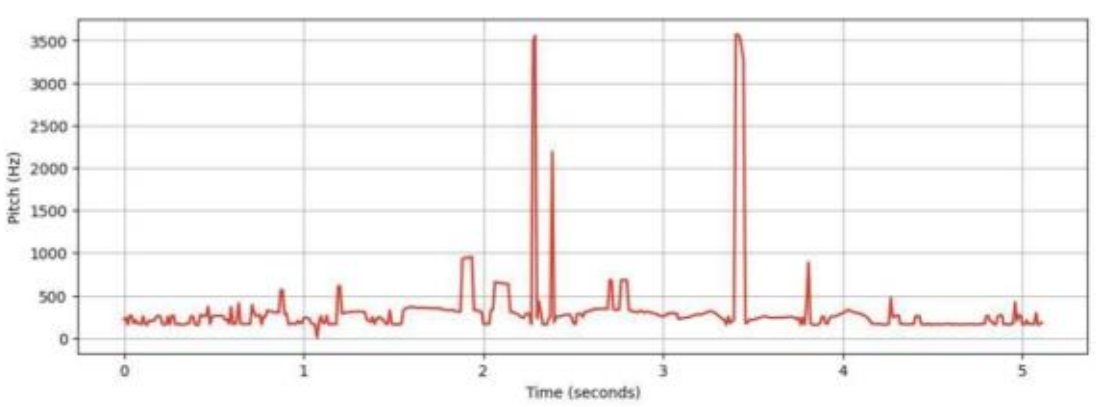

Figure 4: Pitch Tracking

Figure 4 pitch tracking displays the fundamental frequency fluctuations, which dreamily correlate with emotional state(e.g. high pitches with joy or anxiety). This feature extraction confirms the functional requirement of real-time intonation visualization.

The visualization aims to provide a real-time representation of the pitch changes, associating high pitches with active emotions (joyfull), mid-range with neutral emotions, and low pitches with awake emotions as referenced as [45].

As concluded in [45] prosody attributes high pitch levels to emotions, such as joy, anxiety, or fear, because high pitch draws the listener's attention by establishing a contrast. Medium pitch levels account for more neutral attitudes. However, low pitch levels are related to sober emotions: sadness, calmness, or security. Thus, emotions carrying a high level of activity, such as joy or fear, tend to be situated at the top end of the frequency spectrum of the speaker.

This experiment consists of two stages: First, the audio archive is transcribed into text using OpenAI's Whisper model, and its output is directly used as input for the second stage. Subsequently, under the guidance of specific system prompts, the transcribed text is input into the GPT-4o model to summarize the core emotions and analysis. The final text output then contains this summary and a suggested solution. It should be noted that there are costs associated with using GPT-4o.

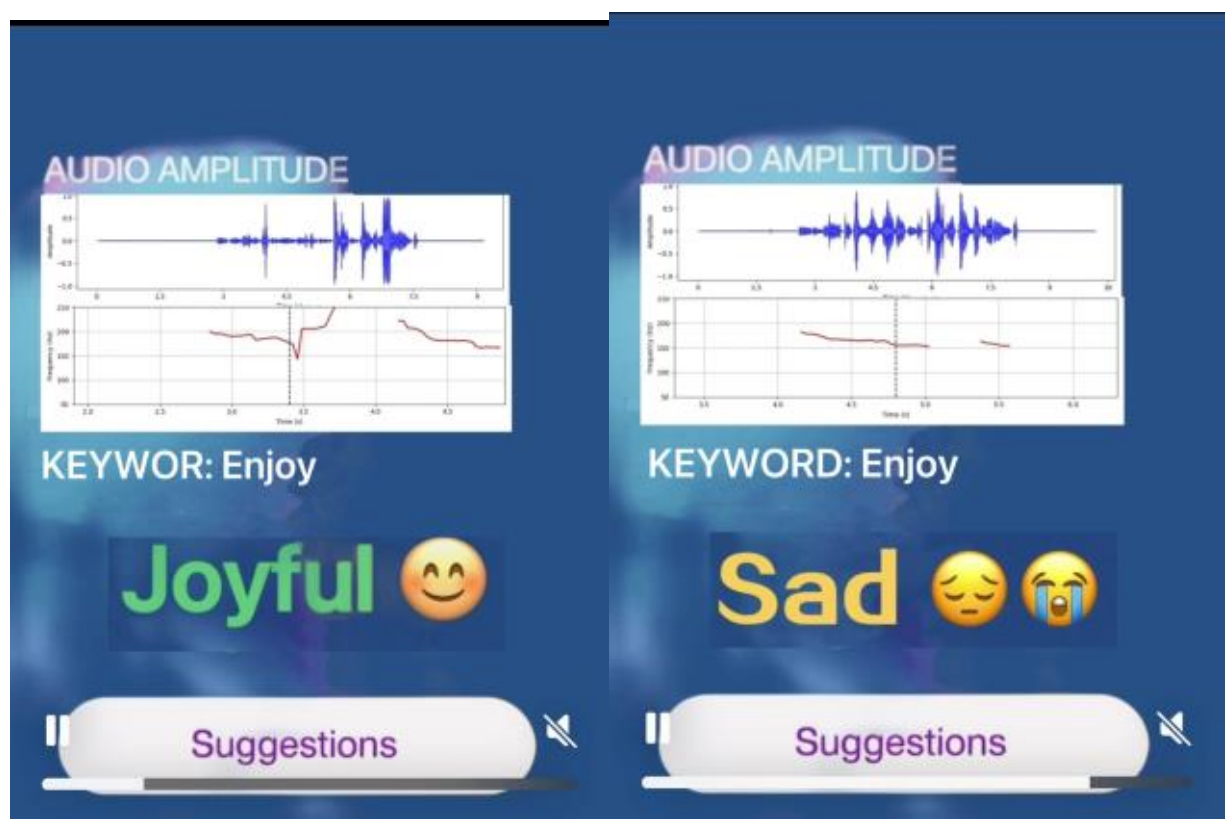

Figure 5: Demo Video

In addition to completing the experiments and demos above, we created a demonstration video Figure 5 using two randomly selected audio samples with the
same content: "I enjoy taking leisurely strolls through the quiet countryside." This video showcases the visual visualization and the effects of different audio and intonation.

Although the text content was the same, the sentence was spoken with two distinct tones. This difference is clearly visualized in the audio amplitude and pitch graphs, despite the consistent keyword tags. The final output confirms this, concluding with "Joyful" for the first sample and "Sad" for the second.

This demonstration video highlights the application's real-time context and low- latency performance. The final conclusion, along with the visualization updates, is delivered just as, or a few seconds after, the sentence ends.

## 6 Conclusion

This paper presents EGI, a novel, lightweight multimodal AI application designed to enhance human emotion verification in agile software development. By integrating real-time ASR, intonation visualization, and lexicon-based keyword prompting, the system provides Scrum Masters with critical emotional insights during team ceremonies. Our experimental evaluation confirms the technical viability of the ensemble architecture, with the Whisper-based ASR module achieving a 10% WER on emotional speech datasets.

The research demonstrates that a modular, microservices-based approach allows for scalable emotion tracking—balancing free, local real-time feedback with on-demand advanced AI recommendations via GPT-4o. While the current iteration successfully bridges a critical gap in Scrum communication, future work will focus on fine-tuning the ASR models with domain specific datasets, implementing advanced noise reduction, and exploring local LLM integration to reduce latency in AI-driven suggestions. Ultimately, this work provides a foundational framework for applying multimodal AI to foster team self-awareness and communication efficiency in industry standard agile environments.

## ACKNOWLEDGMENTS

The author would like to express their sincere gratitude to Dr. Peter Bloodsworth for his invaluable guidance, academic supervision, and continuous support throughout the development of this research at the University of Oxford. The author also thanks the agile practitioners whose feedback informed the system requirements and the broader academic community for the open-source datasets and models that made this work possible.